\documentclass[sigconf]{acmart}

\usepackage{color, colortbl}

\AtBeginDocument{%
  }

\begin{document}


\title{HIINT: Historical, Intra- and Inter- personal Dynamics Modeling with Cross-person Memory Transformer}

\author{Yubin Kim}
\email{ybkim95@media.mit.edu}
\affiliation{
  \institution{MIT Media Lab}
  \city{Cambridge}
  \state{MA}
  \country{USA}
}

\author{Dong Won Lee}
\email{dongwonl@media.mit.edu}
\affiliation{
  \institution{MIT Media Lab}
  \city{Cambridge}
  \state{MA}
  \country{USA}
}

\author{Paul Pu Liang}
\email{pliang@cs.cmu.edu}
\affiliation{
  \institution{Carnegie Mellon University}
  \city{Pittsburgh}
  \state{PA}
  \country{USA}
}

\author{Sharifa Algohwinem}
\email{sharifah@media.mit.edu}
\affiliation{
  \institution{MIT Media Lab}
  \city{Cambridge}
  \state{MA}
  \country{USA}
}

\author{Cynthia Breazeal}
\email{cynthiab@media.mit.edu}
\affiliation{
  \institution{MIT Media Lab}
  \city{Cambridge}
  \state{MA}
  \country{USA}
}

\author{Hae Won Park}
\email{haewon@media.mit.edu}
\affiliation{
  \institution{MIT Media Lab}
  \city{Cambridge}
  \state{MA}
  \country{USA}
}

\renewcommand{\shortauthors}{Kim et al.}

\begin{abstract}
  Accurately modeling affect dynamics, which refers to the changes and fluctuations in emotions and affective displays during human conversations, is crucial for understanding human interactions. By analyzing affect dynamics, we can gain insights into how people communicate, respond to different situations, and form relationships. However, modeling affect dynamics is challenging due to contextual factors, such as the complex and nuanced nature of interpersonal relationships, the situation, and other factors that influence affective displays. To address this challenge, we propose a Cross-person Memory Transformer (CPM-T) framework which is able to explicitly model affective dynamics (intrapersonal and interpersonal influences) by identifying verbal and non-verbal cues, and with a large language model to utilize the pre-trained knowledge and perform verbal reasoning. The CPM-T framework maintains memory modules to store and update the contexts within the conversation window, enabling the model to capture dependencies between earlier and later parts of a conversation. Additionally, our framework employs cross-modal attention to effectively align information from multi-modalities and leverage cross-person attention to align behaviors in multi-party interactions. We evaluate the effectiveness and generalizability of our approach on three publicly available datasets for joint engagement, rapport, and human beliefs prediction tasks. Remarkably, the CPM-T framework outperforms baseline models in average F1-scores by up to 7.3\%, 9.3\%, and 2.0\% respectively. Finally, we demonstrate the importance of each component in the framework via ablation studies with respect to multimodal temporal behavior. 
  
\end{abstract}

\maketitle

\section{Introduction}

In social interactions, individuals rely on a combination of cues to perceive and comprehend the affective states of others, enabling them to gain insights into the contextual aspects of the interaction \cite{doi:10.1146/annurev-psych-010418-103145, knapp2013nonverbal, DBLP:journals/corr/abs-1901-02884, picard2000affective}. This process of understanding is influenced by multiple factors, such as the specific situation at hand, the nature of the relationship between the individuals involved, and the observer's own emotions, experiences, and expectations. Additionally, in the context of multi-party interactions, challenges arise from both inter-personal influences, which involve dynamics among participants, and intra-personal influences, which pertain to individual contributions within the group \cite{dowell2018applying, lee2023multipart}. These challenges highlight the need for a comprehensive approach that considers the lasting impact of affects, the influences of individual and group dynamics, and the contextual nuances within social interactions.

\begin{figure*}[h]
  \centering
  \includegraphics[width=\textwidth]{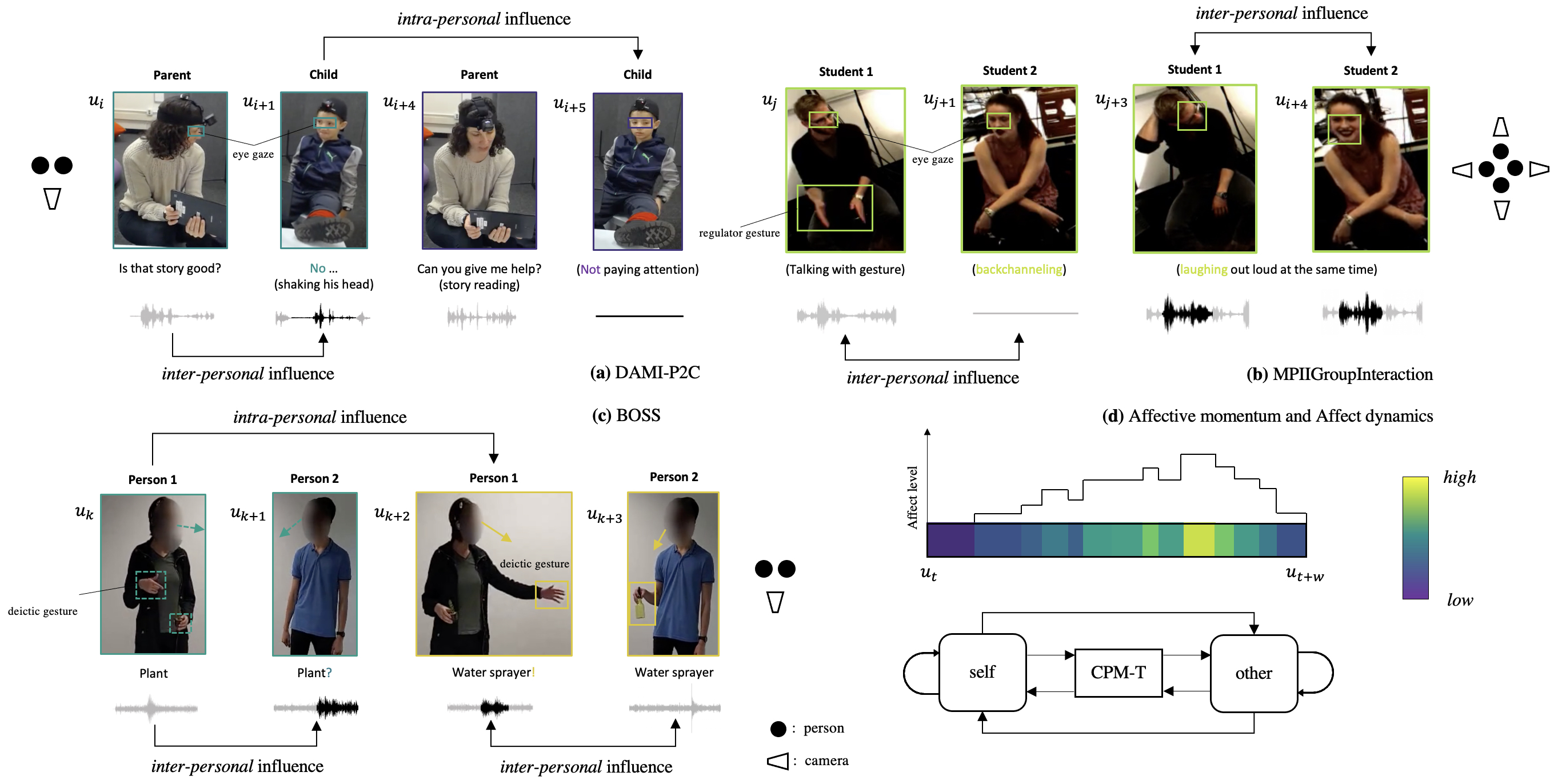}
  \caption{Best viewed zoomed in and in color. Interactive conversation scenarios from (a) DAMI-P2C, (b) MPII, and (c) BOSS datasets. For each conversation, \textit{intrapersonal influence} and \textit{interpersonal influence} are evident with the affective displays and this overall affective momentum with the self and interpersonal influence dynamics are depicted in (d).}
  \Description{A woman and a girl in white dresses sit in an open car.}
  \label{fig:affect_dynamics}
\end{figure*}

To address the challenges posed by affect dynamics in interactive conversations, we propose the Crossperson Memory Transformer (CPM-T) for modeling affect dynamics in interactive conversations. Our model incorporates a cross-modal transformer \cite{tsai2019multimodal} to obtain fused representations of multiple modality features extracted by modality-specific backbones. Additionally, we leverage cross-person attention \cite{lee2023multipart} to capture the influences of intrapersonal and interpersonal factors by encoding verbal and nonverbal cue features. The model also includes a memory network that allows for the retention of past interactions and utilizes the reasoning capabilities of a large language model to guide the interpretation of verbal cues. Given intrapersonal and interpersonal inputs from multiple modalities, CPM-T applies cross-modal and cross-person attention to encode nonverbal representations. This encoding process, guided by verbal reasoning from a large language model and supported by the memory modules, enables the model to autoregressively output an embedding that encapsulates contextualized information about the affective dynamics and interactions in the ongoing conversation. By capturing the momentum of affective states and the complex dependencies between individuals and their historical context, CPM-T enables a deeper understanding of the interplay between verbal and nonverbal cues in social interactions.

To evaluate the effectiveness of our proposed approach, we selected three complex social and affective dynamics tasks: joint engagement, rapport, and human belief prediction from DAMI-P2C \cite{9784429}, MPIIGroupInteraction \cite{mueller18_iui}, and BOSS \cite{duan2022boss} datasets, which involve long-term dependencies influenced by various intra- and inter-personal dynamics. These tasks share commonalities that involve interpreting nonverbal cues, understanding social dynamics and context, possessing empathy and theory of mind, aligning communication, demonstrating cognitive flexibility, and engaging in collaborative problem-solving. By addressing these aspects, our proposed approach aims to enhance social cognition, communication skills, and interpersonal understanding in human interactions.

To summarize, the main contributions of our work are as follows:
\begin{enumerate}
\item We propose Crossperson Memory Transformer (CPM-T), a novel transformer-based model which combines the concept of Cross-person Attention (CPA) and Memory (Slot) Attention for capturing the intra- and inter- personal relationship between pairs of people that lies in long-term dependencies with multi-modal streams. 
\item We utilize the Large Language Model (LLM)'s reasoning as verbal context which provides guidance for nonverbal cues through the memory network to improve the model's performance. 
\item We successfully integrate the proposed model into the joint engagement, rapport, and belief dynamics prediction task on three publicly available datasets. Experiments, ablation studies, and qualitative analysis support the effectiveness of our model and open up new possibilities for improving social human-robot interaction in various settings. 
\end{enumerate}

\begin{figure*}[h]
  \centering
  \includegraphics[width=16cm]{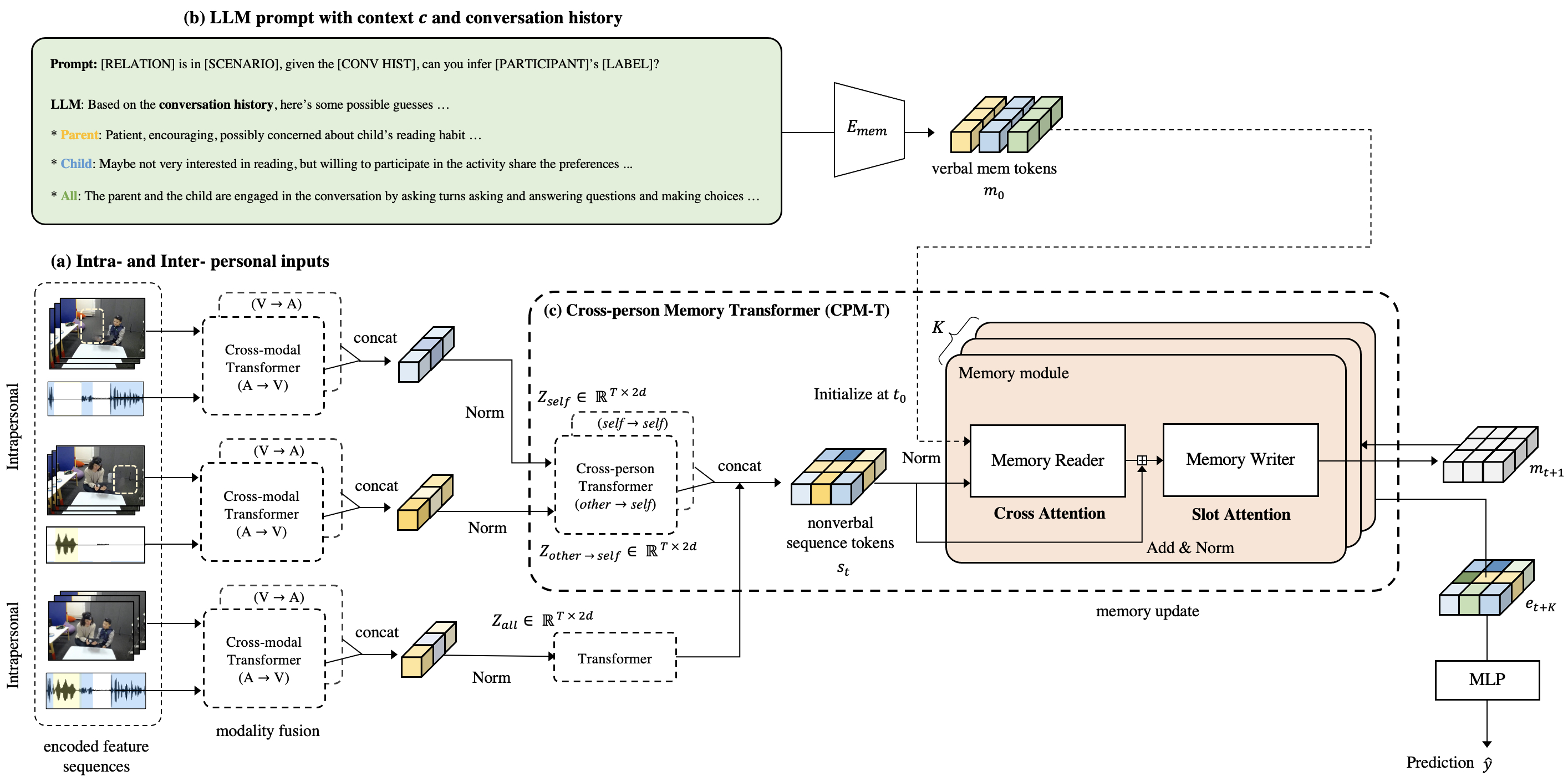}
  \caption{Schematic visualization of the proposed method. (a) while there exist different combinations of modality inputs, we exemplary consider the case of joint engagement prediction between parent and child using audio, video, and conversation history inputs here. The original input is separated into intrapersonal and interpersonal inputs. (b) we utilize verbal cues to guide nonverbal cues by initializing the memory bank with the verbal context and feeding nonverbal segments with this memory to the network (for memory encoder, we borrow the encoder part of the memformer network \cite{wu2022memformer} and utilize the outputs as verbal contexts). (c) nonverbal cues are split into $K$ segments along the temporal axis and iteratively processed through the memory network to update the memory and output the encoded representation in the last layer. Crossperson Attention (CPA) is used to explicitly model the affect dynamics and the mechanism is described in section \ref{subsec:cpm-t}.}
  \Description{A woman and a girl in white dresses sit in an open car.}
  \label{fig:model}
\end{figure*}
%

\section{Related Works}

\subsection{Memory Networks}
Memory networks have gained considerable attention due to their ability to capture and leverage contextual information for understanding and modeling affective experiences. One specific challenge involves effectively capturing the temporal dynamics inherent in affective experiences. In response, researchers have investigated the use of memory networks for modeling interactive conversational memory networks in tasks such as emotion recognition \cite{koval2021emotional, lee2021compm, shen2021dialogxl}, sentiment analysis \cite{xu2019multi}, and emotion flip reasoning \cite{kumar2022discovering}. These models significantly enhance the comprehension and prediction of affective states within real-world interactions. Despite the promising potential of memory networks to address these challenges, further research is necessary to enhance their scalability, interpretability, and generalization capabilities in the context of affective communication tasks. 

\subsection{Modeling Interactive Conversations}
The field of conversational modeling has increasingly recognized the importance of incorporating affect dynamics into understanding human interactions. Specifically, \cite{DBLP:journals/corr/abs-2109-09487} proposes DyadFormer, multi-modal transformer architecture to model individual and interpersonal features in dyadic interactions for personality prediction. \cite{lee2023multipart} present MultiPar-T, a transformer-based model that can capture the contingent behavior in a multi-party setting by conducting an engagement prediction task. \cite{ng2022learning} models interactional communication in dyadic interaction by autoregressively outputting multiple possibilities of corresponding listener motion. \cite{hazarika2018icon} proposes a multimodal emotion detection framework that extracts multimodal features from conversational videos and hierarchically models the self- and inter- speaker emotional influences into global memories. Among these prior works, only a few studies have addressed the challenges of modeling affect dynamics in more complex conversational tasks; joint engagement, rapport, and belief dynamics prediction. Previous models have been limited in their ability to capture the nuances of human interactions by only recognizing a limited range of affective states and contextual cues. Furthermore, they have not fully accounted for the complexities of multimodal features. By addressing these limitations, our proposed model represents a comprehensive approach to modeling affect dynamics in interactive conversations, building on prior work on affect dynamics, multimodal features, and complex conversational tasks. By recognizing a broader range of affective states and contextual cues, our model can capture the nuances of human interactions and enable more accurate modeling of affective dynamics.

\subsection{Language Models as Multimodal Guides}

Language Models (LMs) have proven to be powerful tools in various domains, including affective computing. They have been successfully applied in guiding other modalities for video segmentation \cite{liang2023local}, context-aware prompting \cite{DBLP:journals/corr/abs-2112-01518}, and image classification \cite{yang2023language}. These studies highlight the potential of combining verbal context with nonverbal context to enhance the understanding and generation of nonverbal behaviors in affective computing. By leveraging the capabilities of LMs, we can effectively bridge the gap between verbal and nonverbal cues, enabling a more comprehensive and nuanced understanding of affective dynamics in human interactions. This integration allows us to capture the interplay between verbal and nonverbal expressions, fusing nonverbal behaviors that align with the given verbal context.

In our work, we extend the application of LMs to the modeling of affect dynamics in interactive conversations. By utilizing a large language model as a guiding source for nonverbal cues, we aim to enhance the performance of our proposed Crossperson Memory Transformer (CPM-T) framework in capturing the intricate relationship between verbal and nonverbal aspects of affective communication. Through the integration of verbal context provided by LMs, CPM-T can generate more contextually relevant and emotionally expressive nonverbal behaviors, thereby improving the overall fidelity and naturalness of affective communication modeling. The utilization of LMs in the affective computing domain not only enriches our understanding of human interactions but also opens up new possibilities for applications in social robotics, virtual agents, and human-computer interaction. By effectively combining verbal and nonverbal cues, we can create more engaging and empathetic systems that can better understand and respond to users' affective states and needs.

\section{Methods}

In this section, we describe our proposed Crossperson Memory Transformer (CPM-T) (Figure \ref{fig:model}). At the high level, CPM-T takes fused multi-modal representation from each person using Cross-modal Transformer and utilize Crossperson Attention (CPA) to discover the self and interpersonal influences. Next, we utilize Memory (Slot) Attention modules to incorporate an external dynamic memory to encode and retrieve past information. In Section \ref{subsec:verbal memory}, and \ref{subsec:cpm-t}, we present in details about the ingredients of the CPM-T architecture (see Figure \ref{fig:model}) and explain the importance of each component. 

\subsection{Problem Statement} 
Consider a set of video-audio pairs $\mathcal{D} = \{ (X, y) \}$ where $X$ is the audio and video input and $y \in Y$, is the label from a set of $N$ classes. We extract the features of all audio and video clips in $\mathcal{D}$ as ${X_a = E_{a}(X[a]) \in {\mathbb{R}^{T_{a} \times {d_a}}}}$ and ${X_v = E_{v}({X[v]}) \in {\mathbb{R}^{T_{a} \times {d_a}}}}$, respectively (for certain models, we add extra modalities such as pose ${p}$ and text ${t}$ along with audio and video). Given the task-specific concepts $C = \{c_{1}, c_2, ..., c_N \}$ and the task, we generate a set of reasoning sentences $s = LLM(C)$  and feed these sentences to the memory encoder $E_{mem}$, to generate verbal memory $mem = E_{mem}(s)$. Combined with the Cross-person Memory Transformer model, it produces a prediction, $\hat{y} = f\big(g(X, mem)\big)$, in which $g$ is the CPM-T model, $f$ is the MLP layer, $X$ is the sequence tokens and $mem$ is the memory tokens.

\subsection{Intrapersonal Input Separation} 

\begin{figure}
  \includegraphics[width=8cm]{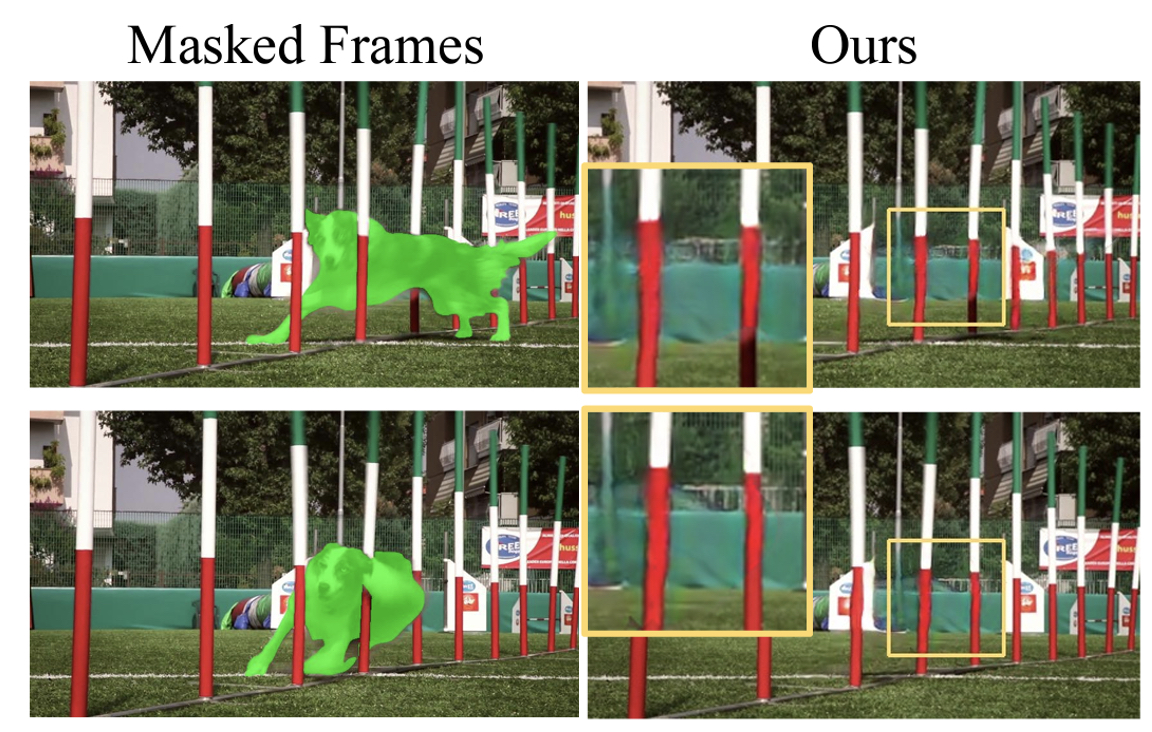}
  \caption{Video Inpainting. The corresponding modules work in an end-to-end manner. A qualitative visualization of the approach is shown in the figure  (image borrowed from \cite{liCvpr22vInpainting}).}
  \label{fig:video_inpainting}
\end{figure}
\begin{figure}
  \includegraphics[width=8cm]{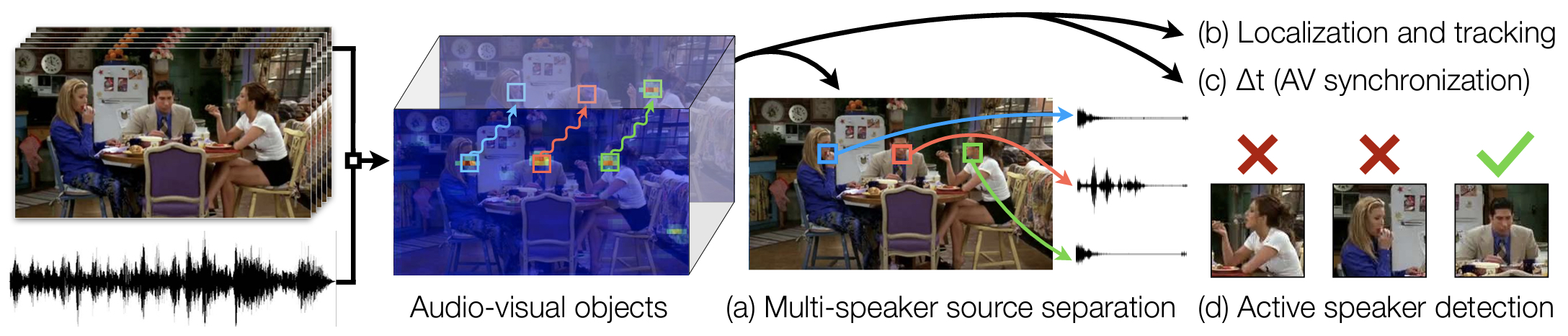}
  \caption{Speaker Diarization. The model learns through self-supervision to represent a video as a set of discrete audio-visual objects. This model groups a scene into object instances and represents each one with a feature embedding (image borrowed from \cite{Afouras20b}).}
  \label{fig:avobjects}
\end{figure}

In Figure \ref{fig:model}, we display the individuals are separated from the original videos, and this process was done by 1) video inpainting (Figure \ref{fig:video_inpainting}) and 2) speaker diarization (Figure \ref{fig:avobjects}). 

\paragraph{\textbf{-Video Inpainting}} 
For video inpainting, we use a flow-guided video inpainting model, $E^{2}FGVI$ \cite{liCvpr22vInpainting} which can handle videos with arbitrary resolution. This model exhibits a strong ability to generalize effectively to higher resolutions, as demonstrated by experimental results and validated performance metrics such as PSNR and SSIM. The video inpainting process can be divided into three interconnected stages. Firstly, flow completion is performed to estimate the missing optical flow fields in corrupted regions, as the absence of flow information in those areas can impact subsequent processes. Secondly, pixel propagation is employed to fill the holes in corrupted videos by bi-directionally propagating pixels from visible areas, leveraging the completed optical flow as a guide. Finally, content hallucination takes place, where the remaining missing regions are generated through the use of a pre-trained image inpainting network.

\paragraph{\textbf{-Speaker Diarization}} For speaker diarization, we use 
the model that uses attention to localize and group sound sources, and optical flow to aggregate information overtime which is presented in \cite{Afouras20b}. The performance of the model has been validated through four downstream speech-oriented tasks: (a) multi speaker sound source separation, (b) localizing and tracking speakers, (c) correcting misaligned audio-visual data, and (d) active speaker detection which showed the effectiveness of the model's learned audio-visual object embeddings. 

\label{appendix:individual_separation}

\subsection{Verbal Memory from LLM Reasoning}

In this paper, we harness the power of LLMs to facilitate the extraction and integration of nonverbal cues within the realm of affective dynamics in interactive conversations. Specifically, we use OpenAI's Chat-GPT, which possesses advanced capabilities to understand and generate human-like text based on given prompts. By leveraging the reasoning abilities of LLMs, we can tap into their deep understanding of language and utilize their contextual comprehension to enhance our understanding of affective states in conversational settings. We prompt the model to generate a set of reasoning about the joint affective states, taking into account the context $c$ and the conversation history within the designated window $w$. The choice of window size determines the amount of context we consider for analysis. For example, for the DAMI-P2C dataset, we provide the following contexts to make a final formatted prompt:

\begin{itemize}
  \item the type of relationship: parent-child
  \item the type of activity they are doing: story reading
  \item the conversation history within window size $w$
  \item the label we want to predict: joint engagement
  \item the entities: parent, child, and both
\end{itemize}

The prompt can be obtained by filling the contextual information to the pre-defined format (see Figure \ref{fig:model}) and we collect responses for 1) parent, 2) child, and 3) dyad by inputting the prompt to the large language model. the encoder part from the memformer \cite{wu2022memformer} which includes the memory reading and writing operations to generate verbal context. This verbal context is used to initialize the memory part of crossperson memory network which takes the nonverbal cue segments as input and updates the nonverbal context guided by the verbal context.

By incorporating LLMs into our methodology, we open up possibilities for more nuanced and insightful analysis of affective dynamics. The integration of nonverbal cues with verbal memory allows us to capture a more holistic view of affective experiences in social interactions. This approach offers valuable insights into the complex interplay between verbal and nonverbal communication, contributing to a deeper understanding of affective inertia and its temporal aspects.

\label{subsec:verbal memory}

\subsection{Crossperson Memory Network (CPM-T)}
In order to successfully address the complex affect dynamics taking place in interactive conversations, we must properly represent each person’s individual nonverbal cues, address self and interpersonal influences, then take into account the long-range dependencies that last over time. 

\paragraph{\textbf{Affect Dynamics Encoding: Cross-person Attention (CPA)}} To explicitly model the self- and interpersonal influences between pairs of people, we utilized Cross-person Attention (CPA), proposed in \cite{lee2023multipart}. This method states that given a pair of people, the target person $self$'s behavior is \emph{contingent on} person $other$'s behavior if person $self$'s behavior was likely to be influenced by person $other$'s behavior ($other$$\rightarrow$$self$). Hence, for the target person $self$, and another person $other$, we utilize the multi-modal representations $Z_{self}$, $Z_{other } \in$ $\mathbb{R}^{T \ \times \  2d}$ obtained from the Cross-modal Transformer, where $T$ is the sequence length and $d$ is the projected feature dimension. Following how Multimodal Transformer \cite{tsai2019multimodal} calculated the cross-modal attention, the cross-person attention can be calculated similarly as below: 

\begin{equation}
\begin{aligned}
& \textstyle{\mathrm{CPA}_{other \rightarrow \text self}\left(Z_{other}, Z_{self}\right)=\operatorname{softmax}\left(\frac{Q_{other} K_{self}^{\top}}{\sqrt{d_x}}\right) V_{self}} \\
& =\operatorname{softmax}\left(\frac{Z_{other} W_{Q_{other}}\left(Z_{self} W_{K_{self}}\right)^{\top}}{\sqrt{d_x}}\right) Z_{self} W_{V_{self}} .
\end{aligned}
\end{equation}

$\text{CPA}_{other \rightarrow self}^{m, \text{multi}}$ refers to the multi-headed from person \textit{other} to person \textit{self} at the \textit{m}-th layer. $\text{CPA}_{other \rightarrow self}(Z_{other}, Z_{self})$ outputs an embedding which has captured the person $self$'s behavior contingent on person $other$'s behaviors. Note that depending on the task, we concatenate the different outputs from the Crossperson Transformers (e.g., for DAMI-P2C (child-coordinated joint engagement), we concatenate the outputs only from ${parent \rightarrow child}$ and ${child}$ whereas we concatenate the outputs from ${other \rightarrow self}$, ${self \rightarrow other}$ and ${self}$ for rest of datasets (rapport, human belief dynamics)).

\paragraph{\textbf{Dynamic Memory Update: Memory Slot Attention}} To encode and retain important past context, we utilize the external dynamic memory method presented in \cite{wu2022memformer}. The model interactively encodes and retrieves the information from memory in a recurrent way by conducting memory read and write operations. At each timestep $t$, we have $M_t = [m_t^0, m_t^1, ..., m_t^k]$. For each slot in the batch, they keep separate memory representations by working individually. For each segment sequence, the model first read the memory to retain the past important information by using cross-attention.  

\begin{equation}
\begin{aligned}
Q_x, K_M, V_M = xW_Q, M_tW_K, M_tW_K \\
A_{x,M} = \operatorname{MHAttn}(Q_x, K_M) \\
H_x = \operatorname{Softmax} (A_{x,M}) V_M
\end{aligned}
\end{equation}

Here, we project memory slot vectors into keys and values and input sequences into queries and use these queries to attend to all key-value pairs in the memory slots, ultimately resulting in the output of the final hidden states. This enables the model to learn the complex association of the memory. Next, memory writing happens with a slot attention module to update memory information and a forgetting method to clean up unimportant memory information. Memory writing only occurs at the last layer of the encoder and allows high-level contextual representations to be stored in memory. Slot attention happens in this stage where each memory slot only attends to itself and token representations, and this prevents each memory slot to write its own information to other slots directly, as memory slots should be independent of each other.

\begin{equation}
\begin{aligned}
Q_{m^i}, K_{m^i} = m^{i}W_Q, m^{i}W_K \\
K_x, V_x = xW_K, xW_V \\
A_{m^i}^{'} = \operatorname{MHAttn} (A_{m^i}, [K_{m^i}; K_x])
\end{aligned}
\end{equation}

Each slot is separately projected into queries and keys. The segment token representations are projected into keys and values. Slot attention means that each memory slot can only attend to itself and the token representations. Thus, each memory slot cannot write its own information to other slots directly, as memory slots should not be interfering with each other. Finally, after the attention scores are calculated, the raw attention weights are divided by the temperature $\tau$, and the next timestep's memory is collected with attention:

\begin{equation}
\begin{aligned}
A_{m^i} = {{exp(A_i^{'}/\tau)} \over {\sum}_j exp(A_j^{'} / \tau)} \\
m_{t+1}^i = \operatorname{Softmax}(A_{x,M}) [m_t^i; V_x] 
\end{aligned}
\end{equation}

For the details of how the memory read and write operation works, we encourage the readers to refer to Appendix C. 
\label{subsec:cpm-t}

\begin{table*}
  \centering
  \renewcommand{\arraystretch}{1.2}
  \begin{tabular}{p{1.6cm}|c|c| c c c c c c}
    \hline
    \centering{Dataset} & Model & Modality & \multicolumn{3}{c}{All Joint Engagement Classes} & 
    \multicolumn{1}{c}{Low Eng} & \multicolumn{1}{c}{Mid Eng} & \multicolumn{1}{c}{High Eng} \\
    \cline{4-9}
    & &  & Accuracy & Weighted F1 & Macro F1 & F1 & F1 & F1 \\
    \hline
    & I3D & \textit{v} & $0.508 \pm 0.02$ & $ 0.548\pm 0.04$ & $0.416 \pm 0.02$ & $0.232 \pm 0.08$ & $0.290 \pm 0.15$ & $0.724 \pm 0.02$ \\ 
    & MulT &  \textit{a}+\textit{v}+\textit{t} & $0.564 \pm 0.02$ & $0.587 \pm 0.04$ & $0.414 \pm 0.02$ & $0.250 \pm 0.03$ & $0.273 \pm 0.06$ & $0.719 \pm 0.02$ \\ 
    \centering{DAMI-P2C} & DyadFormer  &  \textit{a}+\textit{v} & $0.494 \pm 0.01$ & $0.500 \pm 0.01$ & $0.362 \pm 0.03$ & $0.186 \pm 0.04$ & $0.250 \pm 0.08$ & $0.651 \pm 0.01$ \\ 
     & Multipar-T & \textit{a}+\textit{v}+\textit{p} & $0.573 \pm 0.01$ & $0.589 \pm 0.01$ & $0.417 \pm 0.01$ & $0.191 \pm 0.01$ & $0.361 \pm 0.02$ & $0.698 \pm 0.01$ \\ 
    \cline{2-9}
    & \cellcolor{gray!20} \textbf{Ours} & \cellcolor{gray!20} \textit{t} $\rightarrow$ \textit{a}+\textit{v} & \cellcolor{gray!20} $\textbf{0.634} \pm \textbf{0.00}$ & \cellcolor{gray!20} $\textbf{0.677} \pm \textbf{0.01}^\dagger$ & \cellcolor{gray!20} $\textbf{0.490} \pm \textbf{0.00}^\dagger$ & \cellcolor{gray!20} $\textbf{0.286} \pm \textbf{0.01}$ & \cellcolor{gray!20} $\textbf{0.430} \pm \textbf{0.01}$ & \cellcolor{gray!20} $\textbf{0.755} \pm \textbf{0.00}$  \\ 
    \hline
  \end{tabular}
  \caption{Results and standard deviations for the proposed and baseline models on DAMI-P2C dataset using 3 seeds. In "Modality" column, \textit{$m_i$}+\textit{$m_j$} stands for modality fusion for modality \textit{$m_i$} and \textit{$m_j$}. Eng stands for Engagement. ${m_i \rightarrow m_j+m_k}$ means ${m_j+m_k}$ was guided by ${m_i}$ modality using the memory encoder. For MulT, we provided the sentences from the original transcription as $t$, and for Ours, we used the reasoning output sentences from LLM as $t$ (see Section \ref{subsec:verbal memory} and Appendix B for details). $\dagger$ represents statistical significance over state-of-the-art scores under the paired bootstrap test $(p < 0.05)$ and Bonferroni correction.}
  \label{tab:table1}
\end{table*}

\begin{table*}
  \centering
  \renewcommand{\arraystretch}{1.2}
  \begin{tabular}{p{1.6cm}|c|c| c c c c c c}
    \hline
    \centering{Dataset} & Model & Modality & \multicolumn{3}{c}{All Rapport Classes} & 
    \multicolumn{1}{c}{Low Rap} & \multicolumn{1}{c}{Mid Rap} & \multicolumn{1}{c}{High Rap} \\
    \cline{4-9}
    & &  & Accuracy & Weighted F1 & Macro F1 & F1 & F1 & F1 \\
    \hline
    & I3D & \textit{v} & $ 0.371\pm 0.02$ & $0.333 \pm 0.02$ & $0.278 \pm 0.01$ & $0.000 \pm 0.00$ & $\textbf{0.260} \pm \textbf{0.06}$ & $\textbf{0.575} \pm \textbf{0.05}^\dagger$ \\ 
    & MulT &  \textit{a}+\textit{v} & $ 0.319 \pm 0.01$ & $0.298 \pm 0.02$ & $0.261 \pm 0.01$  & $0.107 \pm 0.01$ & $0.145 \pm 0.05$ & $0.531 \pm 0.01$  \\ 
    \centering{MPII} & DyadFormer  &  \textit{a}+\textit{v} & $0.298 \pm 0.02$ & $0.346 \pm 0.02$ & $0.257 \pm 0.02$ & $0.08 \pm 0.02$ & $0.248 \pm 0.05$ & $0.447 \pm 0.02$ \\ 
     & Multipar-T & \textit{v}+\textit{p} & $ 0.494 \pm 0.01 $ & $0.468 \pm 0.01$ & $0.315 \pm 0.01$ & $0.636 \pm 0.02$ & $0.000 \pm 0.00$ & $0.311 \pm 0.04$ \\ 
    \cline{2-9}
    & \cellcolor{gray!20} \textbf{Ours} & \cellcolor{gray!20} \textit{a}+\textit{v} & \cellcolor{gray!20} $ \textbf{0.534} \pm \textbf{0.04} $ & \cellcolor{gray!20} $ \textbf{0.551} \pm \textbf{0.06} $ & \cellcolor{gray!20} $ \textbf{0.408} \pm \textbf{0.01}^\dagger $ & \cellcolor{gray!20} $ \textbf{0.678} \pm \textbf{0.02} $ & \cellcolor{gray!20} $ 0.01 \pm 0.00 $ & \cellcolor{gray!20} $ 0.537 \pm 0.02 $ \\ 
    \hline
  \end{tabular}

  \begin{tabular}{p{1.6cm}|c|c| c c c c c c}
    \hline
    \centering{Dataset} & Model & Modality & \multicolumn{3}{c}{All Human Belief Classes} & 
    \multicolumn{1}{c}{No Comm} & \multicolumn{1}{c}{Attn Follw} & \multicolumn{1}{c}{Joint Attn} \\
    \cline{4-9}
    & &  & Accuracy & Weighted F1 & Macro F1 & F1 & F1 & F1 \\
    \hline
    & I3D & \textit{v} & $0.211 \pm 0.02$ & $0.282 \pm 0.01$ & $0.272 \pm 0.02$ & $\textbf{0.204} \pm \textbf{0.12}$ & $\textbf{0.304} \pm \textbf{0.10}^\dagger$ & $0.307 \pm 0.10$ \\ 
    & MulT &  \textit{a}+\textit{v} & $0.545 \pm 0.02$ & $\textbf{0.638} \pm \textbf{0.07}^\dagger$ & $0.267 \pm 0.06$ & $0.000 \pm 0.00$ & $0.090 \pm 0.01$ & $0.711 \pm 0.01$ \\ 
    \centering{BOSS} & DyadFormer  &  \textit{a}+\textit{v} & $0.422 \pm 0.03$ & $0.446 \pm 0.02$ & $0.265 \pm 0.01$ & $0.000 \pm 0.01$ & $0.194 \pm 0.02$ & $0.600 \pm 0.01$  \\ 
     & Multipar-T & \textit{v}+\textit{p} & $0.351 \pm 0.03$ & $0.338 \pm 0.03$ & $0.272 \pm 0.01$ & $0.098 \pm 0.02$ & $0.216 \pm 0.03$ & $0.501 \pm 0.05$ \\ 
    \cline{2-9}
     & \cellcolor{gray!20} \textbf{Ours} & \cellcolor{gray!20} \textit{a}+\textit{v} & \cellcolor{gray!20} $ \textbf{0.569} \pm \textbf{0.01}^\dagger $ & \cellcolor{gray!20} $ 0.528 \pm 0.02 $ & \cellcolor{gray!20} $ \textbf{0.292} \pm \textbf{0.00} $ & \cellcolor{gray!20} $ 0.157 \pm 0.01 $ & \cellcolor{gray!20} $ 0.143 \pm 0.02 $ & \cellcolor{gray!20} $ \textbf{0.765} \pm \textbf{0.01} $ \\ 
    \hline
  \end{tabular}

  \caption{Results and standard deviations for the proposed and baseline models on MPIIGroupInteraction and BOSS dataset using 3 seeds. In "Modality" column, \textit{$m_i$}+\textit{$m_j$} stands for modality fusion for modality \textit{$m_i$} and \textit{$m_j$}. Rap, Comm, Follw, and Attn stands for Rapport, Communication, Following, and Attention respectively. For MPII and BOSS datasets, experiments on \textit{t} modality are not reported since it was not provided in the original datasets. $\dagger$ represents statistical significance over state-of-the-art scores under the paired bootstrap test $(p < 0.05)$ and Bonferroni correction.}
  \label{tab:table2}
\end{table*}

\section{Experiments}

In this section, we empirically evaluate the Crossperson Memory Transformer (CPM-T) on three datasets that are frequently used to benchmark human affect communication tasks in prior works \cite{9784429, duan2022boss, mueller18_iui}. Our goal is to compare CPM-T with prior competitive approaches on which almost all prior works employ) and unaligned(which is more challenging, and which 
 CPM-T is generically designed for) multimodal language sequences.

\subsection{Datasets and Evaluation Metrics}
We utilize the DAMI-P2C \cite{9784429}, MPIIGroupInteraction \cite{mueller18_iui}, and BOSS \cite{duan2022boss} (See Appendix \ref{appendix:datasets} for more details) as benchmarks to measure the performance of our proposed method against other baselines. Each task requires the understanding of verbal and nonverbal cues from each person and modeling the affect dynamics to predict the joint labels between pairs of people.  

\paragraph{\textbf{DAMI-P2C}} DAMI-P2C is a corpus of multimodal, multiparty conversational interactions in which participants followed a collaborative parent-child interaction to elicit their joint engagement. The dataset was collected in a study of 34 families, where a parent and a child (3-7 years old) engage in reading storybooks together. From the original five-point ordinal scale [-2,2], we modified the labels to three discrete categories for the classification task; Low, Mid, and High joint engagement. A ten-second window was selected as the fragment interval of target audio-visual recordings for the annotation to capture the long-range context of affect dynamics between pairs of dyads. When annotating the recordings, annotators were instructed to judge whether a given fragment contained the story-related dyadic interaction and filter out those that did not. In total, 16,593 fragments have been utilized with 488.03 ± 123.25 fragments from each family on average. 

\paragraph{\textbf{MPIIGroupInteraction}} MPIIGroupInteraction is a dataset that collected audio-visual non-verbal behavior data and rapport ratings during small group interactions. It consists of 22 group discussions in German, each involving either three or four participants and each lasting about 20 minutes, resulting in a total of more than 440 minutes of audio-visual data. 78 German-speaking participants were recruited from a German university campus, resulting in 12 group interactions with four participants, and 10 interactions with three participants. Since rapport is a subjective feeling that is hard to gauge through any existing equipment, the rapport was self-reported by the participants. Responses were recorded on seven-point Likert scales and we modified the labels into three categories (Low, Mid, and High Rapport) to conduct the classification task with the same model structure across different tasks. Each participant rated each item for other individuals in the group, yielding two rapport scores for each dyad inside the larger group and we evaluate the model's performance by averaging the results in different directions.

\begin{table*}
  \centering
  \renewcommand{\arraystretch}{1.2}
  \begin{tabular}{p{2.4cm}|c|c c c c c c c}
    \hline
    \centering{Dataset} & \centering{Ablation} & \multicolumn{3}{c}{All Joint Engagement Classes} & 
    \multicolumn{1}{c}{Low Eng} & \multicolumn{1}{c}{Mid Eng} & \multicolumn{1}{c}{High Eng} \\
    \cline{3-8}
    & & Accuracy & Weighted F1 & Macro F1 & F1 & F1 & F1 \\
    \hline
    & Ours \textit{w/o} LLM  & $0.618 \pm 0.02$ &  $0.637 \pm 0.02$ & $0.476 \pm 0.01$ & $\textbf{0.328} \pm \textbf{0.02}$ & $0.355 \pm 0.02$ & $0.745 \pm 0.01$ \\ 
    \centering{DAMI-P2C} & Ours \textit{w/o} Memory & $0.613 \pm 0.02$ & $0.651 \pm 0.03$ & $0.478 \pm 0.01$ & $0.327 \pm 0.02$ & $0.392 \pm 0.02$ & $0.727 \pm 0.02$ \\ 
    & Ours \textit{w/o} Individuals & $0.609 \pm 0.02$ & $0.606 \pm 0.03$ & $0.428 \pm 0.02$ & $0.178 \pm 0.04$ & $0.365 \pm 0.02$ & $0.740 \pm 0.01$   \\
    \cline{2-8}
    & \textbf{Ours} & $\textbf{0.634} \pm \textbf{0.00}$ & $\textbf{0.677} \pm \textbf{0.01}$ & $\textbf{0.490} \pm \textbf{0.00}^\dagger$ & $0.286 \pm 0.01$ & $\textbf{0.430} \pm \textbf{0.01}$ & $\textbf{0.755} \pm \textbf{0.00}^\dagger$  \\ 
    \hline
  \end{tabular}
  \caption{Ablation. Effect of ablating key components of our method (CPM-T). We encourage the readers to see Figure \ref{fig:model}. $w/o$ LLM refers to the ablation of verbal memory initialization. $w/o$ Memory refers to the ablation of memory modules in CPM-T but passes through a sequence model to do prediction. $w/o$ Individuals refers to the ablation of video inpainting and speaker dimerization to separate individuals in in-person interaction videos. Results with different combinations of components are displayed, where Ours $w/$ All performs well in general. $\dagger$ represents statistical significance over state-of-the-art scores under the paired bootstrap test $(p < 0.05)$ and Bonferroni correction.}
  \label{tab:ablation1}
\end{table*}

\begin{table*}
  \centering
  \renewcommand{\arraystretch}{1.2}
  \begin{tabular}{p{1.6cm}|c|c c c c c c}
    \hline
    \centering{Dataset} & Ablation & \multicolumn{3}{c}{All Joint Engagement Classes} & 
    \multicolumn{1}{c}{Low Eng} & \multicolumn{1}{c}{Mid Eng} & \multicolumn{1}{c}{High Eng} \\
    \cline{3-8}
    & & Accuracy & Weighted F1 & Macro F1 & F1 & F1 & F1 \\
    \hline
    &  \textit{v} & $0.552 \pm 0.01$ & $0.574 \pm 0.02$ & $0.385 \pm 0.01$ & $0.158 \pm 0.04$ & $0.310 \pm 0.01 $ & $0.687 \pm 0.01$  \\ 
    DAMI-P2C & \textit{a}+\textit{v} & $0.618 \pm 0.02$ &  $0.637 \pm 0.02$ & $0.476 \pm 0.01$ & $\textbf{0.328} \pm \textbf{0.02}$ & $0.355 \pm 0.02$ & $0.745 \pm 0.01$ \\ 
    &  \textit{t} $\rightarrow$ \textit{v} & $0.611 \pm 0.00$ & $0.597 \pm 0.01$ & $0.431 \pm 0.01$ & $0.226 \pm 0.05$ & $0.323 \pm 0.06 $ & $0.743 \pm 0.00$  \\ 
    \cline{2-8}
    & \textit{t} $\rightarrow$ \textit{a}+\textit{v}  & $\textbf{0.634} \pm \textbf{0.00}^\dagger$ & $\textbf{0.677} \pm \textbf{0.01}^\dagger$ & $\textbf{0.490} \pm \textbf{0.00}$ & $0.286 \pm 0.01$ & $\textbf{0.430} \pm \textbf{0.01}$ & $\textbf{0.755} \pm \textbf{0.00}^\dagger$  \\ 
    \hline
  \end{tabular}
  \caption{Ablation. Effect of different combinations of modalities for our method (CPM-T). Again, \textit{$m_i$}+\textit{$m_j$} stands for modality fusion for modality \textit{$m_i$} and \textit{$m_j$}, and ${m_i \rightarrow m_j+m_k}$ means ${m_j+m_k}$ was guided by ${m_i}$ modality using the memory encoder. $\dagger$ represents statistical significance over state-of-the-art scores under the paired bootstrap test $(p < 0.05)$ and Bonferroni correction.}
\label{tab:ablation2}
\end{table*}

\paragraph{\textbf{BOSS}} BOSS is a 3D video dataset compiled from a sequence of social interactions between two individuals in an object-context scenario. The two participants are required to accomplish a collaborative task by inferring and interpreting each other’s beliefs through nonverbal communication. Individuals’ latent mental belief states were annotated, for which ground-truth labels are extremely challenging to obtain. Ten pairs of participants (five pairs of friends and five pairs of strangers) were recruited in 15 distinct contexts to compile the dataset. 900 videos from both the egocentric and third-person perspectives were gathered, totaling 347,490 frames. However, the focus on object matching alone does not capture the rich nonverbal communication that occurs during social interactions. To capture these nonverbal cues and enable a more comprehensive analysis of social interactions, the annotation in the BOSS dataset has been modified in this work to include information on participants' joint attention, attention following, and communication. In detail, we define joint attention, attention following, and no communication using a threshold-based approach. Specifically, we considered an instance of joint attention when the number of matched objects exceeded a threshold of 30. For attention following, we set the threshold to a value between 0 and 30. Finally, we defined no communication as an instance where there were no matched objects which were inspired by \cite{DBLP:journals/corr/abs-2104-02841}.

\subsection{Baselines}

We compare our proposed model with a family of baselines in emotion recognition, action recognition, and personality recognition. We run the latest versions of these models and report their scores on a unified benchmark.  For affect recognition models, we compare CPM-T to MulT \cite{tsai2019multimodal} and MultiPar-T \cite{lee2023multipart}. For the action recognition model, we compare our method with I3D \cite{DBLP:journals/corr/CarreiraZ17}. Finally, for personality recognition model, we compare CPM-T to DyadFormer \cite{DBLP:journals/corr/abs-2109-09487}.

\subsection{Implementation Details} We train our models on 4 NVIDIA GeForce GTX 2080 Ti with different training settings which are described in Appendix A. For all three datasets, we conduct cross-validation by iterating through 0.1 proportion of the groups' data as the test, 0.2 proportion of the other groups' data as the valid, and the rest of the other groups' data as the train set for 3 seeds. Our code can be found in the [Anonymous] and will be shared with the dataset access link through the GitHub repository with a camera-ready version.




\section{Results \& Discussion}

In this section, we discuss the quantitative results of our experiments. We compare our approach CPM-T with state-of-the-art baselines. Then, we discuss the importance of each component in the framework and modality used to train the model through ablation studies. Finally, we leave qualitative analysis in Appendix \ref{appendix:qual_anal} to show different types of memory slots obtained from memory writer and crossmodal attention weights to show the correlation learned from audio-visual inputs. Drawing on prior research \cite{lee2023multipart}, we report the macro, weighted F1-score, and F1-score for every class, as well as an accuracy metric. The macro F1-score, calculated as the unweighted average of per-class F1-scores, holds significant value in our study as it signifies the model's performance across all classes, irrespective of their representation in the dataset.

\paragraph{\textbf{Comparison against baseline models}} 

\begin{figure*}
  \includegraphics[width=\textwidth]{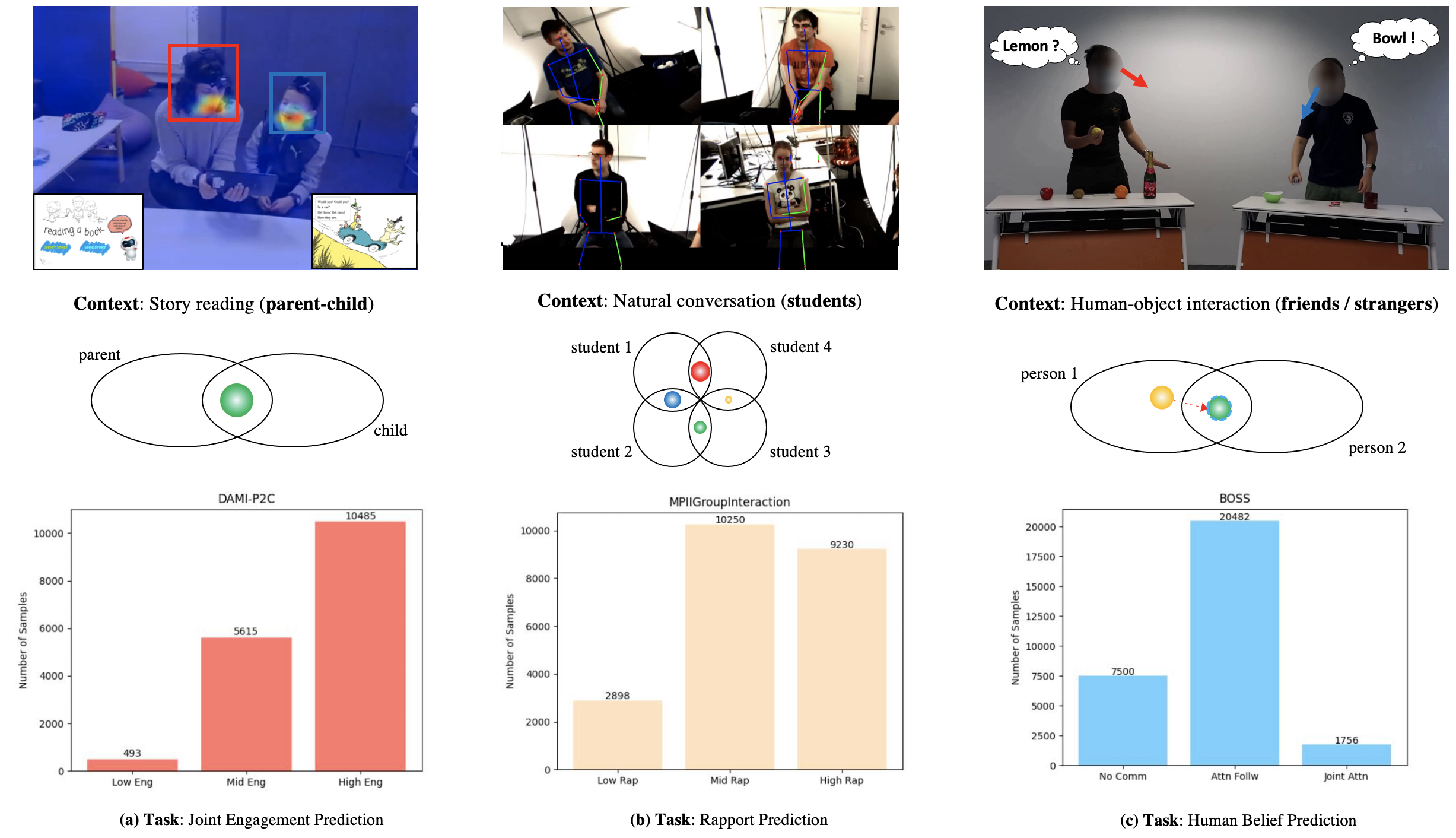}
  \caption{Overview of the affective communication tasks. (a) DAMI-P2C: joint engagement prediction, (b) MPIIGroupInteraction: rapport prediction, (c) BOSS: human belief prediction) which entails the understanding of verbal and nonverbal cues (e.g., facial expression, gesture, gaze, acoustic features, etc).}
  \Description{Enjoying the baseball game from the third-base
  seats. Ichiro Suzuki preparing to bat.}
  \label{fig:teaser}
\end{figure*}

In Table \ref{tab:table1} and \ref{tab:table2}, we evaluated the performance of the proposed model along with baseline models for the task of predicting joint engagement (DAMI-P2C), rapport (MPIIGroupInteraction), and human belief dynamics (BOSS) between dyads. The datasets are highly imbalanced (see Figure. \ref{fig:teaser}) where predicting low engagement, low rapport, and joint attention is challenging. 

For DAMI-P2C, our proposed model, which used audio and video modalities guided by verbal context, achieved the highest performance across all evaluation measures. Moreover, our model achieved a weighted F1-score of 0.677, an improvement of 8.8\%, and the highest macro F1-score of 0.490, an improvement of 7.3\% over the next best-performing model. Particularly, our model outperformed all baselines in predicting the low engagement class, achieving an F1-score of 0.286. This is particularly important given the imbalanced nature of the dataset, with only 493 instances of the low engagement class. Our model's ability to accurately predict low engagement instances could help parents and clinicians identify areas of potential concern and intervene early. In contrast, the I3D model, which only used the video modality, achieved the lowest performance across all evaluation measures. This suggests that the inclusion of other modalities, such as audio and text, can improve the model's ability to predict joint engagement. 

For MPIIGroupInteraction and BOSS, our proposed model achieved the highest performance in accuracy and macro F1-score by using audio and video modality (text modality was not supported in the original dataset). As stated earlier, since the dataset is highly imbalanced, acquiring a high macro F1-score is important. Our model's ability to accurately predict low rapport and joint attention could help teachers or professors to identify the cohesion between students and the mental states of one another. It is also interesting to see that MultiPar-T performed worse compared to the other two datasets and it might be due to the information loss that came from the blurred faces (See Figure \ref{fig:teaser}). In contrast, which only used the video modality, achieved the second-best performance in macro F1-score and we assume this is due to the less meaningful information came from the audio modality (people kept repeating the objects they want to insist).

\paragraph{\textbf{Ablation Studies}}  In order to compare the contribution of each component in our proposed model and the modality we used to train the model with the DAMI-P2C dataset, we performed two ablation studies (See Table \ref{tab:ablation1} and \ref{tab:ablation2}). 

We first systematically removed three components from the full model and compared the resulting performance to the baseline where all components were present. The three components that we removed were the LLM component, the Memory modules, and the separation of individuals from original videos. We measured the accuracy, weighted f1-score, macro f1-score, and f1-scores for each class. Our main interest was the macro f1-score, which provides a better indication of the overall performance of the model when the class distribution is imbalanced. Our results showed that removing any of the three components resulted in a decrease in macro f1-score compared to the baseline. Specifically, removing the LLM component resulted in a decrease of 0.014 in macro f1-score, while removing the Memory modules and the component that separates individuals from in-person interaction videos resulted in decreases of 0.012 and 0.062, respectively. Notably, the full model achieved the highest macro f1-score of 0.490, which was a statistically significant improvement over the baseline. These results suggest that all three components are important for achieving high performance in joint engagement prediction.

In addition to the first ablation study, we conducted another study to investigate the impact of different modality inputs on our model's performance. Specifically, we tested three different input modalities: video only, audio and video, and video guided by verbal memory. We also tested the same input modalities with the addition of verbal memory guidance. Our results show that using both audio and video inputs significantly improved our model's performance, as indicated by a macro f1-score of 0.476, which was significantly higher than using video input only (f1-score of 0.385). Guiding the video input using verbal memory also improved the performance slightly (f1-score of 0.431), but not significantly so. Our findings suggest that using both audio and video inputs is crucial for accurate joint engagement prediction, and that verbal memory guidance can further enhance the video modality's performance.

\section{Conclusion}

In this paper, we presented Crossperson Memory Transformer (CPM-T), a multi-modal multi-party framework for modeling affect dynamics in interactive conversations. Our model capitalizes on modeling contextual information that incorporates self and inter-speaker influences. We accomplish this by using a memory and crossperson transformer. Experiments show that our model outperforms state-of-the-art models on three benchmark datasets. Extensive evaluations and case studies demonstrate the effectiveness of our proposed model. Additionally, the ability to visualize the attentions brings a sense of interpretability to the model, as it allows us to investigate which utterances in the conversational history provide important emotional cues for the current emotional state of the speaker. In the future, we plan to test our model on other relevant affective communication tasks and also explore more into the property of the momentum that comes from second derivatives.

\paragraph{\textbf{Limitations \& Future Works}}
This paper proposes a novel approach, the Cross-person Memory Transformer (CPM-T), which leverages long-range contextual information to predict affect communication tasks between individuals based on verbal and nonverbal cues. However, it is important to note that while the DAMI-P2C dataset used in this study relied on human-generated context, in real-world applications, it will be necessary for the agent to autonomously capture and reason about the context in order to produce appropriate verbal and nonverbal responses. Furthermore, affective momentum, which is a second-order derivative property that arises in the context of affect dynamics, was not explicitly considered in this study. As such, future research should focus on developing models that take into account this property and other higher-order affective phenomena.


\bibliographystyle{ACM-Reference-Format}
\bibliography{samples/icmi_ref}

\newpage
 
\appendix

\section{Datasets}
\subsection{DAMI-P2C}
DAMI-P2C dataset was collected for capturing natural story-reading interactions between a parent and their
child in a lab setting. The dataset consists of five major categories of content (audio and video, sociodemographic profiles, reading behavior features, affect annotations, and person identification and body tracking) necessary to understand the social-emotional behaviors and affective states of parent-child dyads in the co-reading context. This dataset focused on the parent-child co-reading interaction activity, a practice positively associated with both children’s later reading and language outcomes and their interest and enjoyment in reading later in childhood. To capture the parent-child engagement quality, the Joint Engagement Rating Inventory (JERI) \cite{adamson2016joint} was selected, as it quantitates and qualities the caregiver-child interaction during a joint activity where verbal and nonverbal behaviors related to engagement are observed and rated. Specifically, we selected to use Child Coordinated Engagement (CCE) in this work which involves the child’s engagement with the parent
instead of their engagement with the activity. The child’s CCE
will be rated low if the child is engaging in story listening or
reading without attending to the parent and acknowledging
their presence.

\subsection{MPIIGroupInteraction}
The data recording took place in a quiet office in which a larger area was cleared of existing furniture. To capture rich visual information and allow for natural bodily expressions, they used a 4DV camera system to record frame-synchronized video from eight ambient cameras. Specifically, two cameras were placed behind each participant and with a position slightly higher than the head of the participant. During the group forming process, experimenters ensured that participants in the same group did not know each other prior to the study. To  prevent learning effects, every participant took part in only one interaction. To increase engagement, experimenters prepared a list of potential discussion topics and asked each group to choose the topic that was most controversial among group members. Afterward, the experimenter left the room and came back about 20 minutes later to end the discussion. Participants were then asked to complete several questionnaires about the other group members.

\subsection{BOSS}
Participants were instructed to form pairs and stand in front of a table. One table contained a list of contextual objects, and the other table contained a collection of objects that could be selected based on the presented context. Each contextual object had at least two and no more than three possible combinations of object table selections. The set of contextual objects is defined as $o_i^{context} = \{$\textit{Chips, Magazine, Chocolate, Crackers, Sugar, Apple, Wine, Potato, Lemon, Orange, Sardines, TomatoCan, Walnut, Nail, and Plant}$\}$ and the set of objects selected to match these contextual objects is defined as 
$o_i^{select} = \{$\textit{Wine Opener, Knife, Mug, Peeler, Bowl, Scissors, Chips Cap, Marker, Water Spray, Hammer, and Can Opener }$\}$. This experimental design allowed for the investigation of participants' ability to match objects with contextual information, and the BOSS dataset contains the collected data from this task. 
\label{appendix:datasets}

\section{Training Details}
\label{appendix:training_details}

\begin{table}[h]
\centering
\begin{tabular}{|l|c c c|}
\hline
\textbf{} & \textbf{DAMI-P2C} & \textbf{MPII} & \textbf{BOSS} \\
\hline
\centering{Batch Size} & 48 & 48 & 32 \\
\centering{Initial Learning Rate} &  3e-3 & 1e-3 & 2e-3 \\
\cline{2-4}
\centering{Optimizer} &  & AdamW \cite{loshchilov2017decoupled} &  \\
\cline{2-4}
\centering{Behavior Dim} &  & 640 &  \\
\cline{2-4}
\centering{$\#$ of Memory Slots} & 128 & 256 & 256 \\
\centering{$\#$ of Epochs} & 20 & 15 & 15 \\
\cline{2-4}
\centering{Focusing parameter $\gamma$} &  & 10 &  \\
\hline
\end{tabular}
\caption{Hyperparameters of CPM-T model for best performance in various tasks.}
\label{tab:example}
\end{table}

\section{Features}
The features for each modality are extracted using the following tools:

\paragraph{\textbf{-Audio}} We use Vggish \cite{DBLP:journals/corr/HersheyCEGJMPPS16} for extracting low level acoustic features. The VGGish model was pre-trained on AudioSet. The extracted features are from the pre-classification layer after activation. The dimension of the feature tensor is 128. 

\paragraph{\textbf{-Vision}} We use R(2+1)D \cite{DBLP:journals/corr/abs-1711-11248} which the model was pre-trained on Kinetics 400. The model expects to input a stack of 16 RGB frames (112x112) and the dimension of the feature tensor is 512.

\paragraph{\textbf{-Pose}} We use OpenFace \cite{baltrusaitis2018openface} which provides normalized eye gaze direction, location of the head, location
of 3D landmarks, and facial action units with a 128-dimensional vector. For the BOSS dataset, the face part was blurred due to privacy issues. However, we could utilize provided OpenPose \cite{8765346} features which support 25-keypoint body/foot keypoint estimation, including 6-foot key points.
\label{appendix:features}

\section{Memory Modules}
\begin{figure}
  \includegraphics[width=8.5cm]{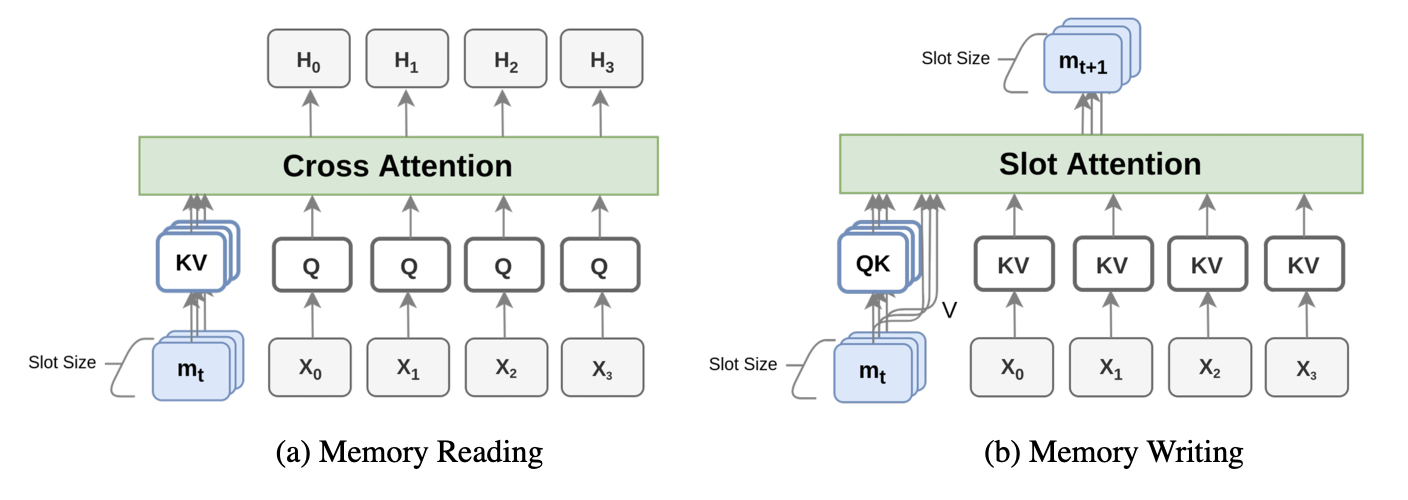}
  \caption{(a) Memory Reading and (b) Memory Writing in memory modules. Image borrowed from \cite{wu2022memformer}.} 
  \label{fig:memory_modules}
\end{figure}

\label{appendix:memory_moduels}

In Figure \ref{fig:memory_modules} (a), the input sequence $x$ undergoes an attention mechanism that encompasses all the memory slots, allowing it to retain historical information. In (b), each memory slot attends over itself and the representations of the input sequence to generate the subsequent memory slot at the next time step. This approach assumes that each memory slot independently stores information and introduces a specific form of sparse attention pattern. In this pattern, each slot in the memory has the ability to attend solely to itself and the outputs of the encoder. The primary objective is to maintain the information within each slot for an extended period throughout the time horizon. By limiting the attention to the slot itself during the writing process, the information contained within that slot remains unchanged in the subsequent timestep. 

In addition, forgetting is an essential aspect of learning since it enables the filtering out of trivial and temporary information, allowing for the retention of more significant and valuable knowledge. In \cite{wu2022memformer}, they propose the utilization of Biased Memory Normalization (BMN), a forgetting mechanism designed specifically for slot memory representations. BMN involves the normalization of memory slots at each step, preventing the memory weights from growing infinitely and ensuring gradient stability over extended periods. To facilitate forgetting of previous information, they introduce a learnable vector bias, $v_{bias}$. The initial state, $v_{bias}^i$, is naturally incorporated after normalization.

\begin{equation}
\begin{aligned}
m_{t+1}^i \leftarrow m_{t+1}^i + v_{bias}^i \\
m_{t+1}^i \leftarrow {{m_{t+1}^i} \over {||m_{t+1}^i||}} \\
m_{0}^i \leftarrow {{v_{bias}^i} \over {||v_{bias}^i||}}
\end{aligned}
\end{equation}

The vector $v_{bias}$ serves as a control mechanism for the rate and direction of forgetting. Introducing $v_{bias}$ to the memory slot, it induces movement along the sphere, resulting in the forgetting of a portion of the stored information. If a memory slot remains unchanged for an extended period, it will eventually reach the terminal state, T, unless new information is injected. The terminal state also serves as the initial state and is subject to learning. The speed of forgetting is determined by the magnitude of $v_{bias}$ and the cosine distance between $m_{t+1}^{'}$ (the updated memory slot) and $v_{bias}$. For instance, if $m_b$ is nearly opposite to the terminal state, it would be challenging to forget its information. On the other hand, if $m_a$ is closer to the terminal state, it becomes easier to forget its information.

\section{Attention Analysis} 
\begin{figure*}[h]
  \centering
  \includegraphics[width=\textwidth]{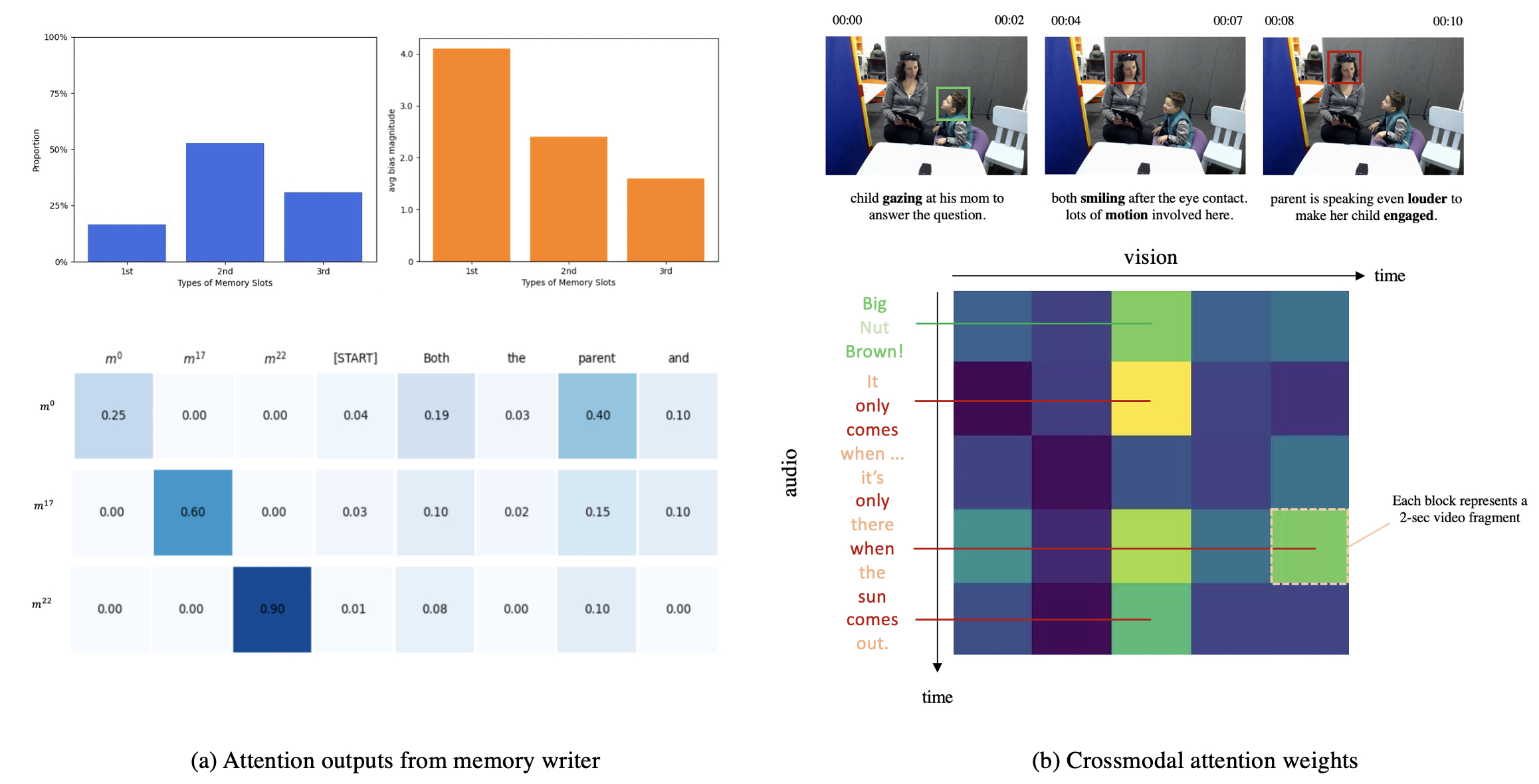}
  \caption{Best viewed zoomed in and in color. (a) Visualization of three different memory slots. We analyze the attention outputs from the memory writer and show the representative memory slots of index 0, 17, and 22 from each category. In the case of the 3rd type (e.g. $m_{22}$) of memory slots, attention focused on themselves, meaning that they were not updating for the current timestep. In contrast, in the case of the 1st type (e.g. $m_{0}$) of memory slots, they completely attended to the input tokens. (b) Visualization of sample crossmodal attention weights from layer 3 of $[V \rightarrow A]$ crossmodal transformer on DAMI-P2C. The attention weights show that the model was able to learn the correlation between audio and vision modalities (especially where people's utterances triggered their facial expressions and bodily gestures).} 
  \label{fig:qual_anal}
\end{figure*}

To demonstrate how the memory network could be used to explain the needs for long-range dependencies, we analyzed the attention outputs from the memory writer following \cite{wu2022memformer}. We empirically categorized the memory slots into three different types and visualized three examples with normalized attention values in Figure \ref{fig:qual_anal} (a). We particularly selected memory slots with indexes of 0,17, and 22. These memory slots represent three types of memories. In 1st type of memory like $m_{22}$, their attention focused on themselves, meaning that they were not updating for the current timestep. This suggests that memory slots can carry information from the distant past. For the second type, the memory slot $m_{17}$ had some partial attention over itself and the rest of the attention over other tokens. This type of memory slot is transformed from the first type of memory slot, and at the current timestep, they aggregate information from other tokens. The third type of memory slot looks like $m_{0}$. It completely attended to the input tokens. In the beginning, nearly all memory slots belong to this type, but later only 5\% to 10\% of the total memory slots account for this type. We also found that the forgetting vector’s bias for $m_{0}$ had a larger magnitude compared to some other slots, suggesting that the information was changing rapidly for this memory slot. 

In addition, to see how crossmodal attention learned the correlation between different modalities \cite{tsai2019multimodal}, we visualize the visualize the attention activation in Figure \ref{fig:qual_anal} (b), which shows an example of a section of the crossmodal attention matrix on layer 3 of the 
$V \rightarrow A$ network (the original matrix has dimension $T_A \times T_V$; the figure shows the attention corresponding to approximately a 10-sec short window of that matrix). We observe that crossmodal attention has learned to attend to meaningful signals across the two modalities. For example, stronger attention is given to the intersection of story-related utterances that tend to trigger the engagement (e.g., “Only Comes”, “Big Brown”) and drastic facial expressions and body gesture change in the video. This observation advocates the advantage of crossmodal transformer over conventional alignment; crossmodal attention enables direct capture of potentially long-range signals, including those off diagonals on the attention matrix \cite{tsai2019multimodal}. 

\label{appendix:qual_anal}
\end{document}